# Global Optimization By Gradient From Hierarchical Score-Matching Spaces

Ming Li
lmcsu@sina.com

**Abstract：** Gradient-based methods are widely used to solve various optimization problems, however, they are either constrained by local optima dilemmas, simple convex constraints, and continuous differentiability requirements, or limited to low-dimensional simple problems. This work solve these limitations and restrictions by unifying all optimization problems with various complex constraints as a general hierarchical optimization objective without constraints, which is optimized by gradient obtained through score matching. The proposed method is verified through simple-constructed and complex-practical experiments. Even more importantly, it reveals the profound connection between global optimization and diffusion based generative modeling.

## 1. Introduction

Global optimization problems represent a critical area of study in science, mathematics, and engineering, aiming to reach the optima and satisfy constraints. These problems often exhibit characteristics such as non-linearity, non-convexity, non-differentiability, complex constraints, and etc. Traditional deterministic methods and stochastic metaheuristic algorithms are the two predominant approaches to handle these challenges.

Stochastic metaheuristic algorithms are still the preferred choice for global optimization. It is mainly divided into several categories: evolutionary-based, swarm-based, physics-based. Evolutionary-based algorithms primarily draw inspiration from Darwin's theory of evolution, utilizing crossover, mutation, and selection within populations to promote survival of the fittest, mainly includes several categories such as Genetic Algorithm (GA) [1], Differential Evolution (DE) [2], Evolutionary Strategies (ES) [3], etc. Swarm-based algorithms is inspired by the collective social behavior of foraging, reproduction, and avoidance of natural enemies, mainly includes several categories such as Particle Swarm Optimization (PSO) [4], Ant Colony Optimization (ACO) [5], etc. Physics-based algorithms primarily involve simulating physical phenomena in nature, with the most renowned algorithm being Simulated Annealing (SA) [6], which is inspired by the metallurgical annealing principle. Recently, evolutionary coding agent based on LLM, such as AlphaEvolve [7], are used to find or improve heuristic algorithms and achieve good performance in some complex optimization problems.

Traditional deterministic algorithms are usually applied to linear, convex, or other simple optimization problems. In this category, original gradient descent and its variants including Quasi-Newton [8], SGD [9], Adam [10], are the most mature and widely used algorithms. However, gradient based methods limited to local optimality, and confined to the field of continuous differentiable problems with simple convex constraints. The only gradient method that does not suffer from the aforementioned issues is the homotopy method, but due to the curse of dimensionality, it is only suitable for low-dimensional simple problems. This article aims to address these issues and limitations.

Our gradient based global optimization theory constructed upon score-matching [11-13], a technique to estimate score, namely, the gradient of the log-density function at the input data point. Score-matching is widely applied to generative models by Sampling with Langevin dynamics [14] or stochastic differential equations (SDE) [15]. Recently, score-matching is gradually replaced by flow based methods [16-20], which estimates velocity, a easier to learn and mutually convertible vector field compared with score.

Score-matching related diffusion models are already applied to solve some optimization problems. DMO [21] solve Gasoline blending scheduling problem, handling constrains by diffusion model and conducting optimization by guidance of objective gradient, while, score-matching here is just used for sampling with constrains and the objective gradient is only locally optimal. DiffUCO [22] solve combinatorial optimization by generative capability of diffusion model similarly.

Different from existing research that treat distribution-sampling by score-matching based diffusion model as optimization method reluctantly, we construct a global hierarchical optimization objective, which is optimized by it's strict gradient that can be obtained from score-matching.

The hierarchical optimization idea in our method is not only incorporated into various metaheuristic algorithms, but also applied to neural network design and gradient optimization. Simple2complex [23] organizes network structure into fractal forms and proves less likely trapping into local minimal that perform gradient optimization in the order of gradually adding high-frequency structures, and it is worth mentioning that the same network was later reinvented by FractalNet [24]. However, this is only an technique specific to the field of neural networks, while the method presented in this paper addresses broader and more general global optimization problems.

Simple2complex, along with other hierarchical optimization methods, share similar idea with homotopy methods, and strictly speaking, the method proposed in this paper is also a homotopy method, but it solves the curse of dimensionality problem inherent in homotopy methods, thus enabling it to solve large-scale complex global optimization problem by rigorous gradient for the first time.

## 2. Motivation

First, let's see how gradient descent do global optimization in ideal situation. Given a minimization optimization function with it's Fourier series form as shown in equation (1):

$$F(x)=\sum_{i=0}^{\infty} a_i \sin(k_i x)+b_i \cos(k_i x) \qquad (1)$$

And it’s low-frequency component sum defined as equation (2):

$$F_t(x)=\left(\sum_{i=0}^{\lfloor t \rfloor} a_i \sin(k_i x)+b_i \cos(k_i x)\right)+(t-\lfloor t \rfloor)\left(a_{\lceil t \rceil} \sin(k_{\lceil t \rceil} x)+b_{\lceil t \rceil} \cos(k_{\lceil t \rceil} x)\right) \qquad (2)$$

We can achieve global optimization for most non-pathological functions as this way: random initialize $x$ in $[x_{min}, x_{max}]$, then do gradient descent by gradient of equation (2) from t=0, when converged then increment t slightly and continue optimization by gradient of equation (2) for the new t, repeat this process until t reach a sufficiently large value that $a_i$ and $b_i$ converge to zero.

For example, we can define a complex fractal objective function by: $k_i = 2^i\pi, a_i =-(-0.7)^i, b_i = (-0.7)^i, b_0 = 0$, which has countless local minimal with domain between 0 and 1, making optimization by traditional gradient descent infeasible, but if we do gradient descent hierarchically as above, we could always achieve global minima as figure 1.

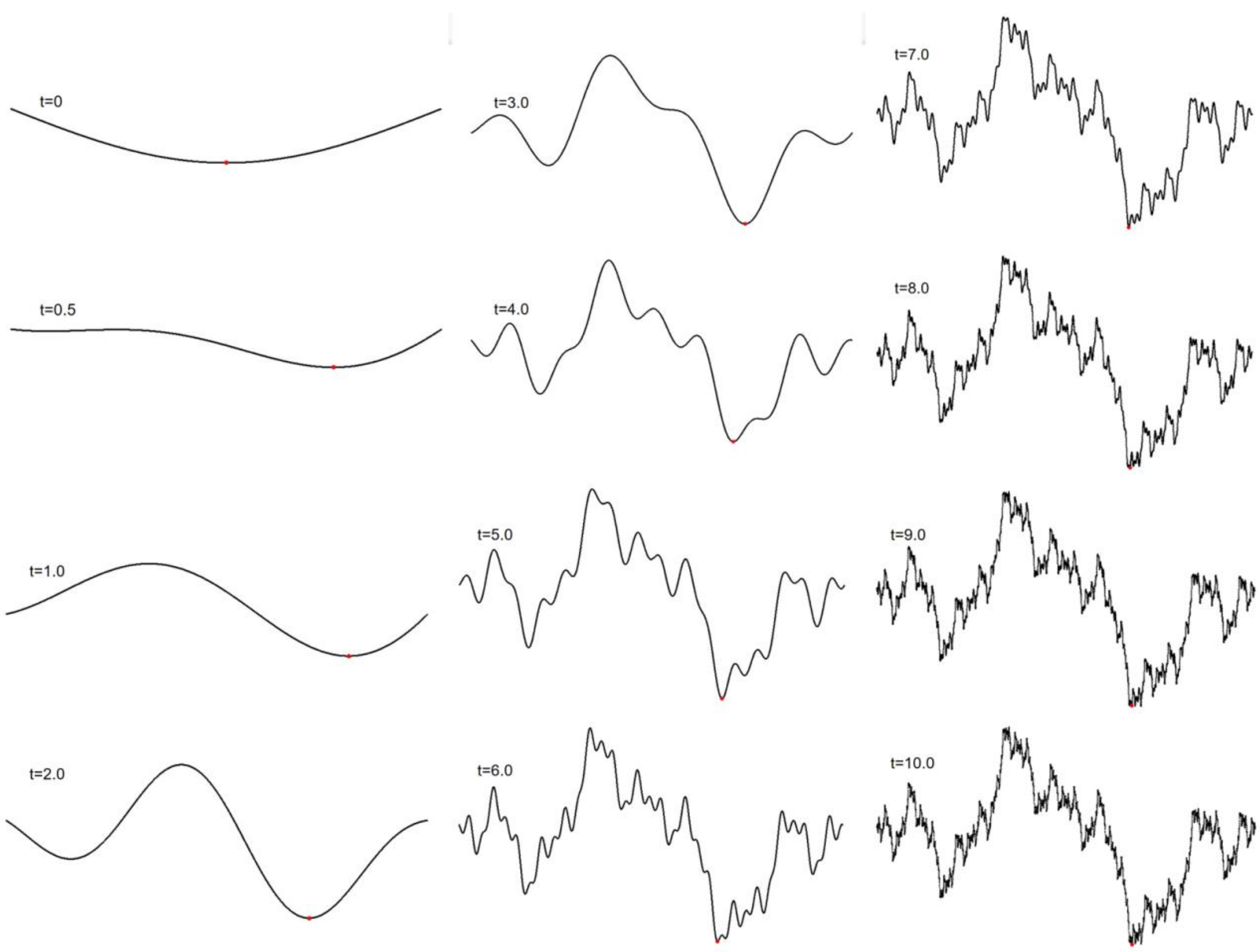

**Figure 1.** $F_t(x)$ growing from a simple convex function to a complex function with countless local minima, the red point is the converged location by gradient descent at current t, as t increments from zero to large value, the red point reach the global minima at last.

Unfortunately, this method is infeasible for most realistic situations, because many problems are non-differentiable or even discrete, even if it is continuously differentiable, obtaining the Fourier series is of exponential complexity for high-dimensional functions, moreover, this method cannot handle complex constraints.

The only general gradient based method capable of solving the aforementioned dilemma is the Gaussian homotopy method: replace the original objective function $f(x)$ with Gaussian convolution as equation (3), and optimize it by gradient as equation (4) using $\sigma$ gradually decreasing to zero. However, this method cannot be widely applied to complex problems due to its dimensionality disaster, to estimate gradient as equation (4) by Monte Carlo sampling, the required number of samples increases exponentially as the dimensionality increases, namely $\mathrm{Num}(\epsilon, n) \sim O(1/\epsilon^n)$, $\epsilon$ is the estimation error bound, furthermore, this method cannot handle complex constraints well.

.

$$H(x) = E_{y \sim N(x, \sigma^2 \cdot I_n)} f(y) \qquad (3)$$

$$\nabla_x H(x) = E_{y \sim N(x, \sigma^2 \cdot I_n)} [f(y)(y - x) / \sigma^2] \qquad (4)$$

This paper proposes the method to address these difficulties in the following order: In section 4, construct a general hierarchical optimization objective handling various complex constrains to replace equation (2) and equation (3), and derive it's gradient and establish connection with score-matching. In section 5, introduce specific methods and solve some technical issues. In section 6, verify through simple-constructed and complex-practical experiments. In section 7, elaborate connection between global optimization and diffusion based generative modeling.

## 3. Flow-based score matching

Let's first recall main ingredients of flow-based score-matching SiT [20], which is necessary for constructing the global optimization theory in section 4.

Flow-based diffusion forward process usually denoted as below:

$$x_t = \alpha_t x + \sigma_t \varepsilon \qquad (5)$$

in which, data x is gradually diffused as $\epsilon \sim N(0, I)$ as t increase from 0 to 1, $\alpha_t, \sigma_t$ are monotonous functions and satisfy $\alpha_0 = \sigma_1 = 1, \alpha_1 = \sigma_0 = 0$.

After flow-matching learning, model output $v(x_t)$ defined as below:

$$v(x_t) = \dot{\alpha}_t E[x \mid x_t] + \dot{\sigma}_t E[\varepsilon \mid x_t] \qquad (6)$$

then, score of $x_t$ defined as equation (7), can be derived from $v(x_t)$ or $E(x|x_t)$ as shown in equation (8):

$$score(x_t) = \nabla_{x_t} \log p(x_t) = \nabla_{x_t} \log \int p(x_t \mid x) p(x) dx \quad (7)$$

$$score(x_t) = \sigma_t^{-1} \frac{\alpha_t v(x_t) - \dot{\alpha}_t x_t}{\dot{\alpha}_t \sigma_t - \alpha_t \dot{\sigma}_t} = \frac{\alpha_t}{\sigma_t^2} (E(x \mid x_t) - \frac{x_t}{\alpha_t}) \qquad (8)$$

In equation (7), $p(x)$ is the data distribution and also sampling distribution for training flow-matching model. $p(x_t|x)$ is the Gaussian distribution decided by equation (5), detailed as equation (9). And the inverse Gaussian distribution $p(x|x_t)$ decided by equation (5) can be derived as shown in equation (10). :

$$p(x_t \mid x) = (2\pi\sigma_t^{\,2})^{-n/2} \exp(-\frac{\| x_t - \alpha_t x \|^2}{2\sigma_t^{\,2}}) \qquad (9)$$

$$p(x \mid x_t) = (2\pi(\sigma_t/\alpha_t)^2)^{-n/2} \exp(-\frac{\| x - x_t/\alpha_t \|^2}{2(\sigma_t/\alpha_t)^2}) \qquad (10)$$

The reverse-time SDE as shown in equation (11) using $v(x_t)$ and $score(x_t)$ is usually used to generate samples from $x_t$, $\overline{W}_t$ is a reverse-time Wiener process, and $w_t > 0$ is an arbitrary time-dependent diffusion coefficient. The reason using the reverse-time SDE generating samples other than the inverse Gaussian of equation (10) is that $p(x)$ always defined with complex constrains and the inverse Gaussian ignore these constrains. There for, from a certain perspective, the reverse-time SDE can be regarded as inverse Gaussian with constrains.

$$dX_t = v(X_t)dt + \frac{1}{2} w_t score(X_t)dt + \sqrt{w_t} d\overline{W}_t \qquad (11)$$

Conversely, if $p(x)$ defined without any constrains, for example, components of x are independent and conform to uniform distribution [-1, 1] continuous or discrete，we can just use equation (10) to generate samples from $x_t$ other than use the model-based reverse-time SDE.

## 4. General global gradient optimization theory

In this section, we first construct a general hierarchical optimization objective handling various complex constrains. Then, derive it's gradient and establish connection with score-matching.

### 4.1 General hierarchical optimization objective

First, define optimization problem with any constrains(discrete, nonlinear, etc) as shown in equation (12), in which $fitness(x) > 0$ and constrains are given by specific problem：

$$\underset{x \in R^*}{\text{Maximize}}\; fitness(x), \quad R^* = \{x \mid x \in R^n \text{ and } x \sim constrains\} \qquad (12)$$

Then, rephrase equation (12) into a equivalent form without constrains as shown in equation (13):

$$\underset{x \in R^n}{\text{Maximize}}\; f(x), \quad f(x) = \begin{cases} fitness(x), & x \in R^* \\ 0, & x \notin R^* \end{cases} \qquad (13)$$

The general hierarchical optimization objective is defined as shown in equation (14), given in the form of product-integral, whose integrated function is a additive-integral of multiplication of $p(x|x_t)$ (equation (10)) and $f(x)$ (equation (13)), and hierarchical scale is defined by t, we will do optimization by gradient ascent as t decrease from 1 to 0 gradually.

$$\text{Maximize } P(x,t) = \tilde{\prod_{x_t \sim R^n}} \left( \int_{\bar{x} \sim R^n} p(\bar{x} \mid x_t) f(\bar{x}) d\bar{x} \right)^{p(x_t \mid x) dx_t} \quad (14)$$

Let's first analysis the meaning of additive-integral in equation (14). $p(\bar{x}|x_t) \sim N(x_t/\alpha_t, (\sigma_t/\alpha_t) \cdot I)$ is an inverse Gaussian distribution, it's quality is given by the additive-integral, the larger the value, the more $\bar{x} \in R^*$ with higher $fitness(\bar{x})$ it will cover.

Then analysis the meaning of product-integral in equation (14). The integrated function is a exponential function, assume the base term is fixed with a specific inverse Gaussian distribution defined by some $x_t$, let's discuss in two scenarios: scenario 1, the base term greater than 1, to maximize the integrated exponential function, we have to adjust x that maximize exponential term $p(x_t|x)$, the optimal solution $x = x_t/\alpha_t$ can be easily found by equation (9), and it's just the center of the inverse Gaussian distribution of base term, therefore, maximizing the whole product-integral means encouraging x close to centers of those inverse Gaussian distribution with high quality; scenario 2, the base term less than 1, similarly, to maximize the integrated exponential function, we have to adjust x that minimize exponential term $p(x_t|x)$, namely, move away from centers of those inverse Gaussian distribution with low quality. And it is worth noting that these two scenario is actually the same thing when you view all the inverse Gaussian distributions as whole together, x will move away from center with quality greater than 1 when all other inverse Gaussian distributions are of higher quality, and x will close to center with quality less than 1 when all others are of lower quality. More rigorously, you can multiply a constant C of infinity or infinitesimal to $f(\bar{x})$ to switch between these two scenario, and you will find C has no effect on the final gradient derived in next section 4.2. Therefore, you don't need to consider the relative magnitude of fitness compared to 1.

**4.2 Gradient by score-matching**

In this section, let's derive gradient of equation (14) and establish it's connection with score-matching.

Comparing equation (9) and (10), it is easy to see that $p(x|x_t) = \alpha_t^n p(x_t|x)$; and let $p(x) = Cf(x)$ as data sampling distribution for score-matching, in which $C$ is normalizing constant, meaning sampling proportional to $f(x)$, namely, sampling proportional to fitness(x) within constrains; and let $C_t = \alpha_t^n/C$; Therefore, the additive-integral in equation (14) can be simplified as below:

$$\int_{\bar{x} \sim R^n} p(\bar{x} \mid x_t) f(\bar{x}) d\bar{x} = \int_{\bar{x} \sim R^n} (\alpha_t^n / C)\, p(x_t \mid \bar{x}) p(\bar{x}) d\bar{x} = C_t p(x_t) \quad (15)$$

Substitute equation (15) and equation (5) into equation (14), then perform simultaneous discretization for $x_t$ and $\varepsilon$, and take the limit, equation (14) can simplified as below:

$$
\begin{aligned}
P(x,t) &= \tilde{\prod_{x_t \sim R^n}} \left(C_t p(x_t)\right)^{p(x_t \mid x) dx_t} \\
&= \exp(\log( \underset{\max \Delta\varepsilon \sim \Delta x_t \to 0}{Lim} \prod_{i=1}^{\infty} \left(C_t p(x_t = \alpha_t x + \sigma_t \varepsilon_i)\right)^{(p(\varepsilon_i)(\Delta\varepsilon / \Delta x_t))\Delta x_t} )) \\
&= \exp( \underset{\max \Delta\varepsilon \to 0}{Lim} \sum_{i=1}^{\infty} \log\left(C_t p(x_t = \alpha_t x + \sigma_t \varepsilon_i)\right) p(\varepsilon_i) \Delta\varepsilon ) \\
&= \exp( \int_{p(\varepsilon) \sim N(0,I)} \log\left(C_t p(x_t = \alpha_t x + \sigma_t \varepsilon)\right) p(\varepsilon) d\varepsilon )
\end{aligned}
$$

Although $\nabla_x P(x, t)$ can not be obtained, but $\nabla_x \log P(x, t)$ can be obtained as below:

$$
\begin{aligned}
\nabla_x \log P(x,t) &= \int_{p(\varepsilon) \sim N(0,I)} (\nabla_{x_t} \log p(x_t = \alpha_t x + \sigma_t \varepsilon) . \nabla_x x_t + \nabla_x \log C_t) \, p(\varepsilon) \, d\varepsilon \\
&= \alpha_t \int_{p(\varepsilon) \sim N(0,I)} score(x_t = \alpha_t x + \sigma_t \varepsilon) \, p(\varepsilon) \, d\varepsilon \qquad (16)
\end{aligned}
$$

To achieve global optimization, the general hierarchical optimization objective (equation (14)) is optimized by gradient ascent (equation (16)) as t gradually decreases from 1 to 0, $\mathrm{score}(x_t)$ in equation (16) can be obtained by score-matching with sampling probability proportional to fitness and within domain of constrains.

Unlike equation (4), estimating equation (16) by Monte Carlo is free of dimensionality disaster. As shown by equation (8), $\mathrm{score}(x_t)$ is gradient-statistics of the distribution center by all data under $N(x_t/\alpha_t, (\sigma_t/\alpha_t)^2 \cdot I)$, Monte Carlo for equation (16) is equivalent to sample some $N(x_t/\alpha_t, (\sigma_t/\alpha_t)^2 \cdot I)$ with $x_t \sim N(\alpha_t x, {\sigma_t}^2 \cdot I)$ and mix them to estimate gradient-statistics of all data under distribution $N(x, 2(\sigma_t/\alpha_t)^2 \cdot I)$. For $\mathrm{score}(x_t)$ is stable and smooth with enough information from all data, Monte Carlo sampling size for equation (16) no need to be large even for high dimentionality.

## 5. Methods

In this part, we will first introduce the basic global optimization process based on above theory in section5.1. Then, in sections 5.2 to 5.7, we address a series of specific technical issues.

### 5.1 Basic global optimization process

Define a exponential decreasing t-sequence $\{t_i | i = 0,1...TN \; and \; t_0 = 1 \; and \; t_{i+1} = \gamma \, t_i\}$, and optimization by gradient ascent is conducted according to the order of this t-sequence, $\gamma$ is a positive constant less than 1.

First, $x^*$ is random initialized in the range of [-1, 1]. For each $t_i$, finetune score-matching model from weights of $t_{i-1}$ using default configuration of SiT but with t fixed to $t_i$ and sampling proportional to fitness within constrains (samples scale and shift to the range of [-1, 1] by default), then $\mathrm{score}(x_t)$ can be obtained from the velocity of model-output by equation (8), and gradient ascent base on equation (16) is used to optimize $x^*$ for current t-scale repeatedly until converged. Finally, after $x^*$ is optimized by all t-scales, scale and shift it to original domain and return as the best solution.

### 5.2 Particularity for the first t-scale

Notice that the gradient of equation (16) for $t_0$ always equals to zero theoretically, so for the first t-scale, it is not necessary to do gradient ascent. But compared to random initialization, we can perform a better initialization at first t-scale. As can be seen from equation (5) that $E(x_t|x_t) = x_t = \alpha_t E(x|x_t) + \sigma_t E(\varepsilon|x_t)$, combine equation(6) and solve the system of equations, we get equation (17) as below:

$$E(x \mid x_t) = \frac{\sigma_t v(x_t) - \dot{\sigma}_t x_t}{\dot{\alpha}_t \sigma_t - \alpha_t \dot{\sigma}_t} \qquad (17)$$

Therefore, we can initialize $x^*$ as below after score-matching at first t-scale:

$$x^* = \underset{\varepsilon \in N(0,I)}{E} E(x \mid x_{t_0} = \varepsilon) \qquad (18)$$

### 5.3 Instability of gradient

Notice that the gradient of equation (16) tends to 0 as t approaches 1 for the reason of $\alpha_t$, and tends to explode as t approaches 0 for the reason of $score(x_t)$. In order to obtain a stable gradient, we transform equation (16) to obtain equation (19) as below:

$$\frac{\sigma_t}{\alpha_t} \nabla_x \log P(x,t) = \int_{p(\varepsilon)\sim N(0,I)} \sigma_t score(x_t = \alpha_t x + \sigma_t \varepsilon) p(\varepsilon)\, d\varepsilon \qquad (19)$$

For $\alpha_t/\sigma_t$ is constant at scale t, we absorb it into learning rate, and $\sigma_t score(x_t)$ can be obtained stably by equation (8), therefore we can use equation (19) as stable gradient replacing the unstable one as equation (16). By this way, optimize by gradient is stable for any t in [0, 1].

### 5.4 Mirror-symmetry Monte Carlo sampling

To obtain the value of equation (16), equation (18) and equation (19), we use Monte Carlo sampling. To eliminate the impact of noises with limit MonteSize as much as possible, we sample $\{\varepsilon\}$ as this way: firstly, sample half of MonteSize by $p(\varepsilon) \sim N(0, I)$; then, for each sampled $\varepsilon$, append $-\varepsilon$ together.

### 5.5 Exponential transformation for fitness

It is found better to do a exponential transformation for fitness before sampling as below.

$$fitness(x) = \exp\left( \frac{fitness(x) - TP_{99}(fitness(x))}{\text{Temp}} \right) \qquad (20)$$

$TP_{99}$ means the top 99th percentile. Temp is used to adjust sampling variance,

can be regarded as weight of optimization objective relative to constrains. In practice, we sample some $\{x\}$, and get $\{fitness(x)\}$ by equation (20), then get sampling probabilities $p = \{fitness(x)/sum(fitness(x)\}$, we adjust $Temp$ to make $std(p) * len(p)$ close to a constant $C_{pstd}$ which take the value of 0.5 by default.

### 5.6 Local efficient score-matching

Without consider efficiency of score-matching, we use random uniform sampling as prior distribution before sampling by fitness. As $t_i$ decreases from 1 to 0, optimization scale evolves from global to local, therefore, we can improve efficiency by using sampling around $x^*$ in a smaller local region as prior distribution.

To sample in local region, define $p_{\theta,t}(\overline{x}|x)$ as equation (21) in which $p(x_t|x)$ is defined by equation (9) and $p_\theta(\overline{x}|x_t)$ is defined by the reverse-time SDE of a specific diffusion model, the specific diffusion model can be a normal SiT model trained in advance.

$$p_{\theta,t}(\overline{x} \mid x) = p_\theta(\overline{x} \mid x_t).p(x_t \mid x) \qquad (21)$$

As discussion in section 3, if components of x are independent and conform to uniform distribution and without constrains, we can just use equation (10) to replace the reverse-time SDE, then, we can simplify $p_{\theta,t}(\overline{x}|x)$ to $p_t(\overline{x}|x)$ as below:

$$p_t(\overline{x} \mid x) = p(\overline{x} \mid x_t).p(x_t \mid x) \sim N(x, 2(\sigma_t/\alpha_t)^2 \cdot I) \qquad (22)$$

Therefore, when do score-matching at $t_i$, we can use $p_{\theta,t_i}(x|x^*)$ or $p_{t_i}(x|x^*)$ to sample in local region around $x^*$ now, but to gain a slightly greater degree of freedom in optimization, better to use $p_{\theta,t_{i-PN}}(x|x^*)$ or $p_{t_{i-PN}}(x|x^*)$ in which PN is a positive integer.

Whether use $p_{\theta,t_{i-PN}}(x|x^*)$ or $p_{t_{i-PN}}(x|x^*)$ as prior distribution, samples can all be regarded as conforming to $p_t(x|x^*) \sim N(x^*, 2(\sigma_t/\alpha_t)^2 \cdot I)$ with or without constrains as figure 2-a, this will bring bias to score-matching that should have used random uniform sampling as prior distribution. To eliminate this bias, we can get a resample weight function $w(x)$ that $w(x).p_t(x|x^*)$ convert distribution in figure 2-a to distribution in figure 2-b, the r in graph take the value to satisfy that the sum probability in region $[x^* - r, x^* + r]$ equals 0.9 by default.

Therefore, one choice to sample as prior without bias is that sample by $p_{\theta,t_{i-PN}}(x|x^*)$ or $p_{t_{i-PN}}(x|x^*)$ first then resample proportional to $w(x)$ again, but this method is only suitable for low-dimensional scenarios, for variance of $w(x)$ will explode soon as the dimensionality increases.

Another choice is based on such an assumption that the effect of resample is equivalent to loss weighting. The method is, after sample by $p_{\theta,t_{i-PN}}(x|x^*)$ or $p_{t_{i-PN}}(x|x^*)$, get $w(x)$ as vector calculated each dimension separately other than get $w(x)$ as scalar, then, when do score-matching, use $w(x)$ as loss-weight for each dimension independently, by this way, we jump the explode problem of resample. If there is no special instruction, we will use this method by default.

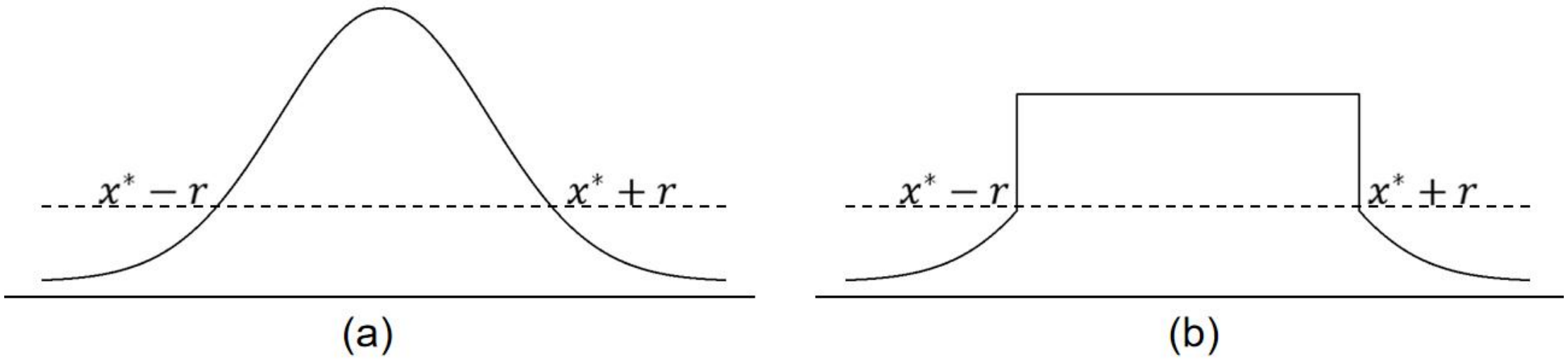

**Figure 2.** the left graph (a) depicts Gaussian prior distribution around $x^*$. the right graph (b) depicts prior distribution without bias except tail region, which is converted from (a) by multiply a resample weight function $w(x)$.

### 5.7 Pathological multi-modal and parallel exploration

Some pathological optimization objective have multiple global optimal solutions, often due to the symmetry of the problem. We solve this problem by parallel exploration with multiple solutions $\{x^*\}$. For each $t_i$, sample for each $x^*$ independently and use all those samples to do score-matching together, after optimize each $x^*$ by gradient ascent independently, merge similar solutions and remove poor ones to ensure that the size of left solutions does not exceed KeepSize, finally, sample ExploreTime new solutions for each $x^*$ by $p_{t_i}(x|x^*)$ and append to $\{x^*\}$ for next t-scale.

## 6. Experiments

In section 6.1-6.2, verify our theory and method by simple constructed experiments. In section 6.3-6.4, it's applied to practical complex problems.

In all experiments, by default, use $TN = 30$ with $\gamma \approx 0.8$ that meet $t_{TN} = 2e-3$, $\mathrm{PN} = 3$; for score-matching of each t-scale, use default configuration of SiT with standard transformer with depth-head-dim 4-4-64, finetune 1k steps with batch size 1024 and learning rate 5e-5; for optimization gradient ascent of each t-scale, use $\mathrm{MonteSize} = 128$ and learning rate 1e-2. In experiments 6.3-6.4, unlike the default, these configurations are used instead: $TN = 67$ with $\gamma \approx 0.9$ that meet $t_{TN} = 1e-3$, $\mathrm{PN} = 4$, transformer with depth-head-dim 8-8-64, finetune 1w steps, and $\mathrm{MonteSize} = 1024$.

### 6.1 Fractal objective function

We use the $F(x)$ with $k_i = 2^i\pi, a_i = -(-0.7)^i, b_i = (-0.7)^i, b_0 = 0$ and domain between 0 and 1 in section 2 to verify our theory and method. Ten replicate experiments with different seeds are conducted, parallel exploration are not used for the objective function is not pathological. As shown in figure 3, the optimized solutions that plotted as red points all fall into the vicinity of the optimal solution.

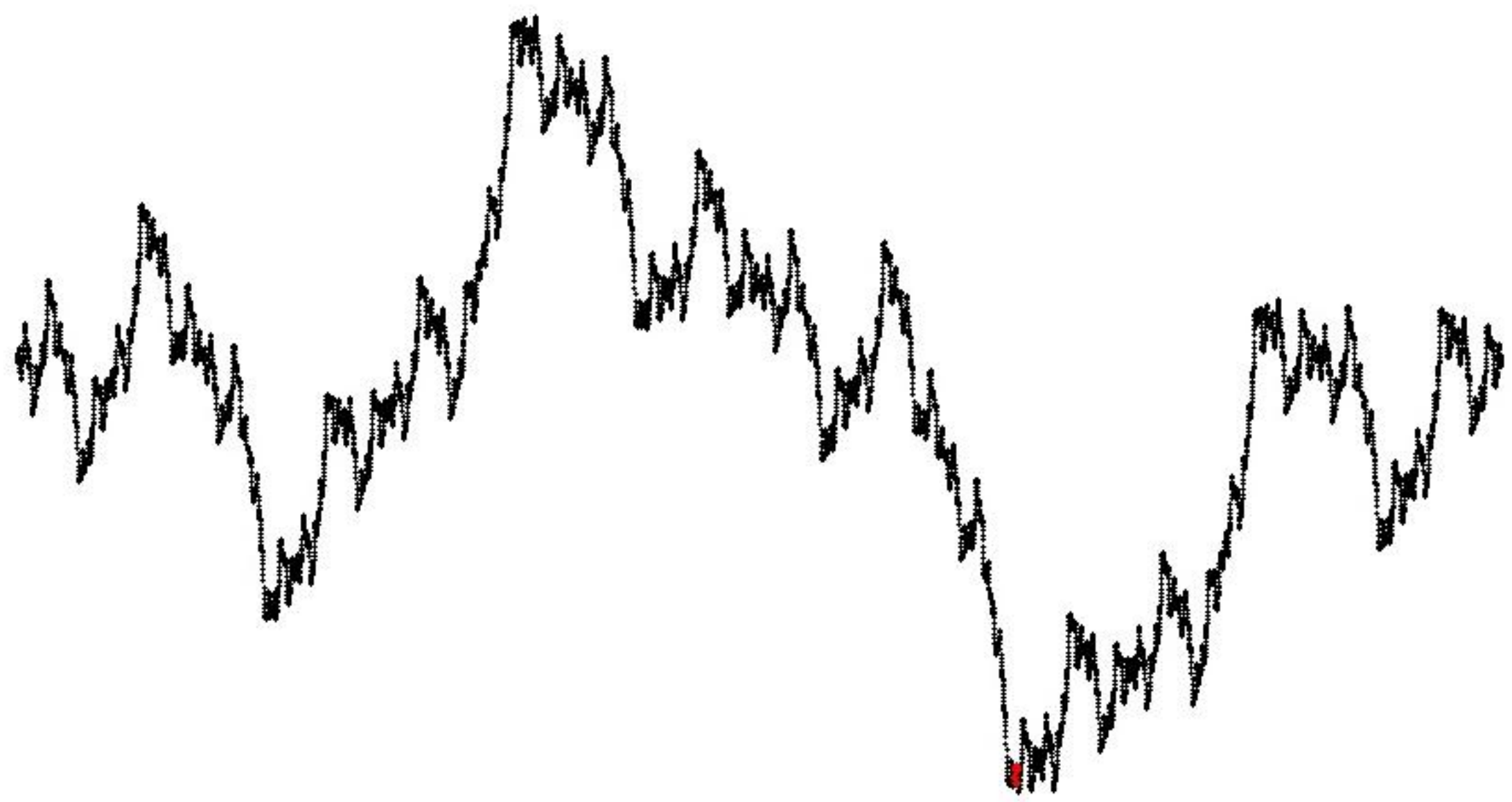

**Figure 3.** the optimization results of 10 experiments for fractal objective function, all optimized solutions marked with red points fall into the vicinity of the optimal solution.

### 6.2 Multi-modal fractal objective function

We construct a pathological multi-modal function as $(\mathrm{F}(\mathrm{x})+\mathrm{F}(1-\mathrm{x}))/2$ with $\mathrm{k_i}=2^{\mathrm{i}}\pi, \mathrm{a_i}=-(-0.7)^{\mathrm{i}}, \mathrm{b_i}=(-0.7)^{\mathrm{i}}, \mathrm{b_0}=0$ and domain between 0 and 1, by this way, the objective function have two optimal solutions. We use parallel exploration with $\mathrm{KeepSize}$ increasing from 4 to 8 and $\mathrm{ExploreTime}$ decreasing from 4 to 2 according to the order of t-sequence. Ten replicate experiments with different seeds are conducted, as shown in figure 4, two solutions closest to the two optimal solutions are selected and plotted as green points for each experiment, and other optimized solutions are plotted as red points. We can see that all experiments succeed to find solution into the vicinity of each of the two optimal solutions.

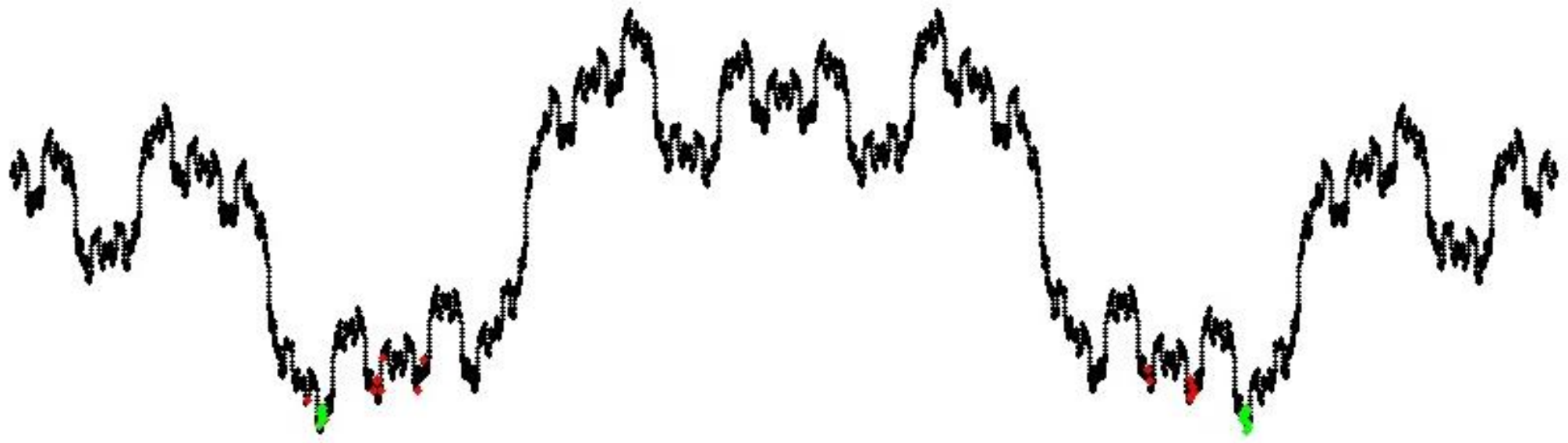

**Figure 4.** the optimization results of 10 experiments for multi-modal fractal objective function, all experiments succeed to find solution (green points) into the vicinity of each of the two optimal solutions.

### 6.3 CEC2017

We try to solve CEC2017 (ndim=30) with our method using transformer with 30 tokens that each embed one dimension. Some objective functions can be solved well, take F42017 (equation (24) with extra rotation and translation) as example, by using local efficient score-matching without parallel exploration, we successfully approach the optimal solution with $x^*$ to a precision of 5e-4 by repeating the optimization process twice, with the second iteration taking place within the domain of the first iteration's result. But when use local efficient score-matching with parallel exploration, all $x^*$ stuck and cannot converge during the optimization process, this indicates that parallel exploration failed for multi-dimension when use local efficient score-matching, it is probably because that training score-matching simultaneously in multiple $x^*$ local domains is mutually influential for multi-dimensional problems, we argue that there may be other better parallel exploration methods, we are considering introducing SDE as a promising candidate.

Another unexpected difficulty we encounter is that scale differences across various dimensions of some objective functions are significant and our method can't handle. Take F12017 (equation (23) with extra rotation and translation) as an example, the scale of first dimension is 1 million times smaller, when $x^*$ is optimized by our method, the first dimension will miss global-optimization with too small gradient signal for starting t-scales, and will get stuck only with local-optimization for ending t-scales. We try to solve this difficulty by adaptive-adjusting scale of each dimension for different $t_i$ like evolutionary strategy, but it still cannot be solved stably, have to left this for future research.

$$F_{12017}(x) = x_1^{\ 2} + 10^6 \sum_{n=2}^{N} x_n^{\ 2}, \quad x \in [-100,100]^N \qquad (23)$$

$$F_{42017}(x) = \sum_{n=1}^{N} (x_n^{\ 2} - 10\cos(2\pi x_n) + 10), \quad x \in [-100,100]^N \qquad (24)$$

We haven't conducted experiments for other objective functions before we can handle the above two difficulties, for that some of them have the problem of scale differences to some extent and some are pathological multi-modal that need use local efficient score-matching with parallel exploration.

### 6.4 Pack n circle in unit square

In this experiment, we try to solve a real complex optimization problem: how to fit n balls into a unit square so that the sum of their radii is maximized, recently, AlphaEvolve [7] improved it from 2.634 to 2.635 for n=26. It is obviously a classical pathological multi-modal problem because of the symmetry of coordinates in terms of left-right, up-down, and transposition, as well as the exchange symmetry between various circles. Considering that parallel exploration still not works well for multi-dimension when use local efficient score-matching, we restrict the domain to $[x^{\wedge} - 0.05, x^{\wedge} + 0.05]$, $x^{\wedge}$ is the optimal solution by AlphaEvolve, in this way, it is no longer a pathological multi-modal problem, but still a highly complex non-convex optimization problem, we conduct this experiment to verify whether our method can solve practical problems, despite some restrictive conditions being imposed.

We use local efficient score-matching without parallel exploration, 26 tokens that each embed one center point coordinates of a circle. We successfully find a different solution $x^*$ with sum radii 2.6350. To achieve greater precision, we repeat the experiment with refined domain $[x^* - 5e - 4, x^* + 5e - 4]$, then obtain a refined solution $x^*$ with sum radii 2.6354, , which is very close to AlphaEvolve's result 2.6358. As shown in figure 5, the solution we found is obviously different compared with AlphaEvolve's.

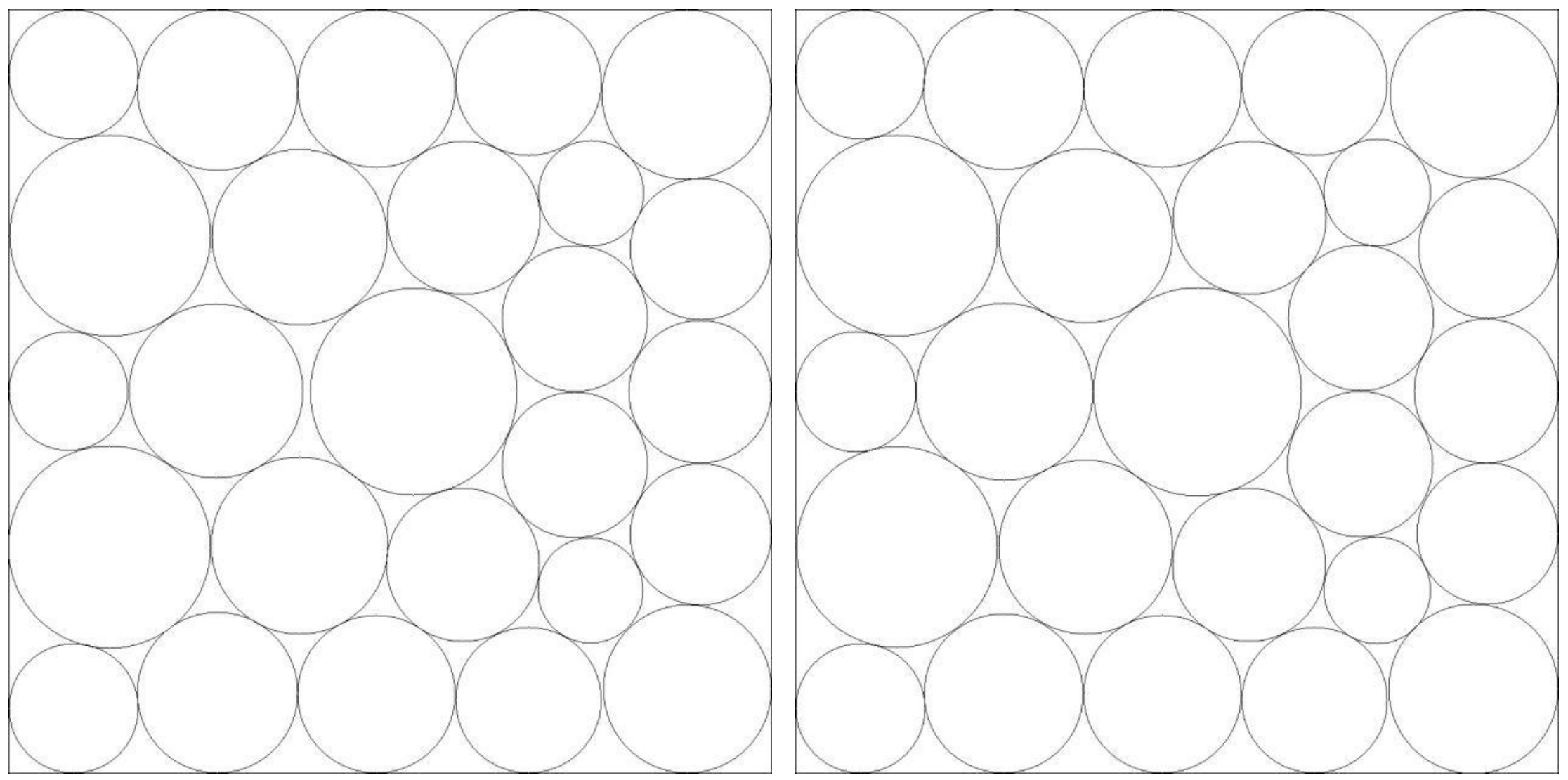

**Figure 5.** the left is AlphaEvolve's result, and the right is the solution we found, there's obvious differences between this two solutions.

## 7. Global optimization based generative modeling

Whether it is a generative model based on diffusion or autoregression, the essence is to fit the probability distribution of the data, and then sample from the distribution, we collectively refer to these models as paradigm "sampling based generative models".

Our global optimization method can be seen as a different paradigm in generative modeling that fit the probability distribution of the data similarly but with sampling-by-fitness, and then generate the best sample with highest probability by global optimization. We call this paradigm "global optimization based generative models".

When applying this new paradigm in practice, some obstacles may be encountered. Take LLM as example, which is a non-pathological problem without obvious symmetry and scale difference. Firstly, We need do score-matching in adequate continuous spaces for our global optimization, but diffusion-based LLM is still dominated by the form based on discrete masking [25-27]. Besides, inference efficiency will become an important consideration that we have to finetune score-matching for every t-scale of every prompt, a potential solution is to train on all t-scales and all prompts simultaneously without using local efficient score-matching. Finally, sampling by fitness may be a problem, for there is only one sample for each prompt at most of the time, one way is treating it as single-point distribution, another way is to train a LLM as domain model and another LLM as fitness model in advance that we can sample from domain model and get fitness.

All in all, we reveal the profound connection between global optimization and diffusion based generative modeling theoretically and believe global optimization based generative modeling is a very promising direction. Due to resource constraints, we will conduct relevant practice and verification in future research.

## 8. Conclusion

In this paper, we propose a global optimization theory and method that purely based on gradient obtained by score-matching, solving the curse of dimensionality problem of homotopy methods, enabling it to solve large-scale complex global optimization problem by rigorous gradient for the first time. The proposed method is verified by simple-constructed and complex-practical experiments. To be frank, there are still two technical issues that remain unresolved: scale differences across various dimensions and parallel exploration for multi-dimension, we welcome interested researchers to work together to solve these problems. We also reveal the profound connection between global optimization and diffusion based generative modeling, anticipating that this will become a new paradigm for generative modeling.